\documentclass[sigconf]{acmart}
\usepackage{multicol}
\usepackage{multirow}
\usepackage{makecell}
\usepackage{subfigure}
\usepackage{amsmath}
\usepackage{algorithmic}            
\usepackage{algorithm}
\usepackage{balance}

\AtBeginDocument{%
  }

\copyrightyear{2022} 
\acmYear{2022} 
\setcopyright{acmcopyright}\acmConference[SIGIR '22]{Proceedings of the 45th International ACM SIGIR Conference on Research and Development in Information Retrieval}{July 11--15, 2022}{Madrid, Spain}
\acmBooktitle{Proceedings of the 45th International ACM SIGIR Conference on Research and Development in Information Retrieval (SIGIR '22), July 11--15, 2022, Madrid, Spain}
\acmPrice{15.00}
\acmDOI{10.1145/3477495.3531811}
\acmISBN{978-1-4503-8732-3/22/07}

\begin{document}
\fancyhead{}

\title{Understanding Long Programming Languages with Structure-Aware Sparse Attention}


\author{Tingting Liu}
\affiliation{%
  \institution{East China Normal University}
  \streetaddress{No. 3663, Zhongshan North Road}
  \city{Shanghai}
  \country{China}
  \postcode{200062}
}
\email{ttliu@stu.ecnu.edu.cn}

\author{Chengyu Wang}
\affiliation{%
  \institution{Alibaba Group}
  \streetaddress{No. 1122, Xiangxiang Street, Yuhang District}
  \city{Hangzhou}
  \state{Zhejiang}
  \country{China}}
\email{chengyu.wcy@alibaba-inc.com}

\author{Cen Chen}
\affiliation{%
  \institution{East China Normal University}
  \city{Shanghai}
  \country{China}
}
\email{cenchen@dase.ecnu.edu.cn}

\author{Ming Gao}
\authornote{Corresponding author.}
\affiliation{%
 \institution{East China Normal University}
 \city{Shanghai}
 \country{China}}
\email{mgao@dase.ecnu.edu.cn}

\author{Aoying Zhou}
\affiliation{%
  \institution{East China Normal University}
  \city{Shanghai}
  \country{China}}
\email{ayzhou@dase.ecnu.edu.cn}

\renewcommand{\shortauthors}{T. Liu et al.}


\begin{abstract}
Programming-based Pre-trained Language Models (PPLMs) such as CodeBERT have achieved great success in many downstream code-related tasks. Since the memory and computational complexity of self-attention in the Transformer grow quadratically with the sequence length, PPLMs typically limit the code length to 512. However, codes in real-world applications are generally long, such as code searches, which cannot be processed efficiently by existing PPLMs. To solve this problem, in this paper, we present SASA, a Structure-Aware Sparse Attention mechanism, which reduces the complexity and improves performance for long code understanding tasks. The key components in SASA are top-$k$ sparse attention and Abstract Syntax Tree (AST)-based structure-aware attention. With top-$k$ sparse attention, the most crucial attention relation can be obtained with a lower computational cost.
As the code structure represents the logic of the code statements, which is a complement to the code sequence characteristics, we further introduce AST structures into attention.
Extensive experiments on CodeXGLUE tasks show that SASA achieves better performance than the competing baselines. 
\end{abstract}




\maketitle

\begin{figure*}
\centering
\includegraphics[width=0.9\textwidth]{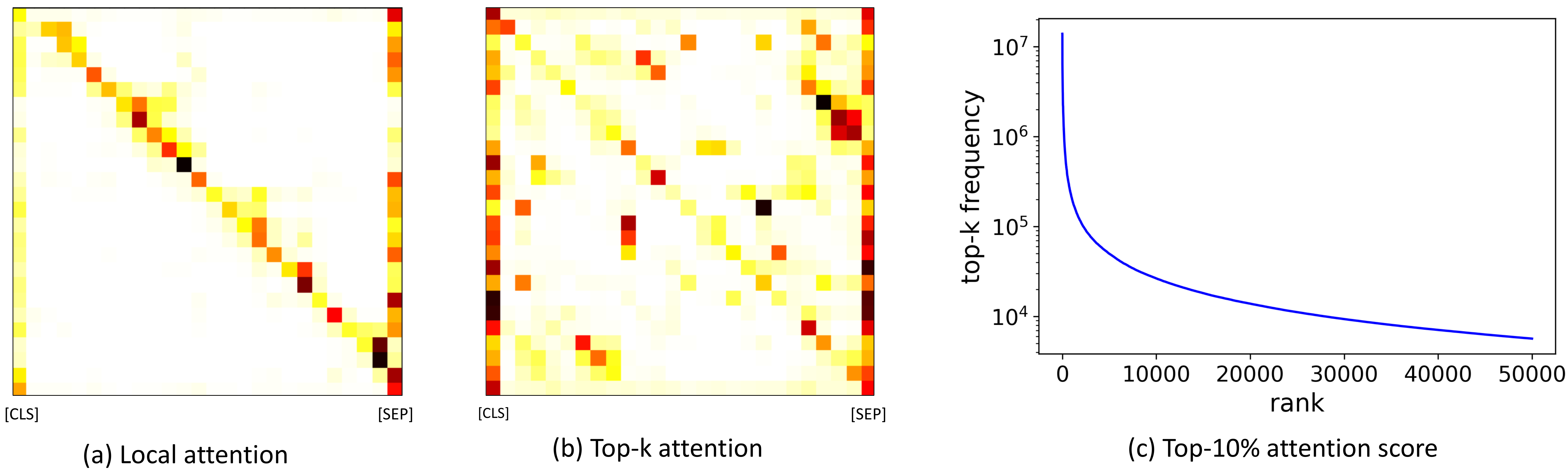}
\caption{Attention patterns in the pre-trained model.
The results show that the attention scores among tokens in long codes present a ``long-tail'' phenomenon.
} 
\label{attention_pattern}
\end{figure*}

\section{Introduction}
Transformer-based pre-trained language models such as BERT~\cite{DBLP:conf/naacl/DevlinCLT19}, RoBERTa~\cite{DBLP:journals/corr/abs-1907-11692}, and ELMo~\cite{DBLP:conf/naacl/PetersNIGCLZ18} have achieved great success in Natural Language Processing (NLP) tasks. These models are usually pre-trained on a large amount of unlabeled data and fine-tuned on downstream tasks. Similarly, existing Program-based Pre-trained Language Models (PPLMs) such as CodeBERT~\cite{DBLP:conf/emnlp/FengGTDFGS0LJZ20}, GraphCodeBERT~\cite{DBLP:conf/iclr/GuoRLFT0ZDSFTDC21} and PLBART~\cite{DBLP:conf/naacl/AhmadCRC21} have significantly improved the performance of code-related tasks such as code clone detection \cite{DBLP:conf/icse/SajnaniSSRL16}, code search~\cite{DBLP:journals/corr/abs-1909-09436}, and code summarization~\cite{DBLP:conf/acl/IyerKCZ16}.

Despite the success, since the memory and computational complexity of self-attention in the Transformer~\cite{DBLP:conf/nips/VaswaniSPUJGKP17} grow quadratically with the sequence length, PPLMs typically limit the code length to 512. Although 512 is suitable for some tasks,
codes in industrial scenarios are generally long and cannot be processed efficiently by existing PPLMs. 
There are two solutions to this case.
One is to truncate the long sequence to 512, which undoubtedly has information loss. The other is to process the full sequence, but this leads to the $O(n^2)$ complexity in both time and memory consumption w.r.t. the sequence length $n$.
Recently, several approaches have been developed to process long sequences efficiently in NLP. For example, sparse attention methods~\cite{DBLP:journals/corr/abs-1904-10509,DBLP:conf/nips/ZaheerGDAAOPRWY20,DBLP:conf/emnlp/QiuMLYW020} sparsify the attention matrix by limiting the attention field to a fixed range.
Therefore, these methods can handle longer sequences using similar hardware. However, even though programs and natural languages are both token sequences, we observe that these sparse attention methods do not work well for long codes, which have different syntactical and structural features from natural languages.

In this paper, we present SASA, a Structure-Aware Sparse Attention mechanism, which reduces the complexity and improves performance for long code understanding tasks. SASA consists of two key components, i.e., top-$k$ sparse attention and Abstract Syntax Tree\footnote{\url{https://en.wikipedia.org/wiki/Abstract_syntax_tree}} (AST)-based structure-aware attention. 
Specifically, top-k sparse attention allows each token to attend to only $k$ tokens with the highest attention scores. 
As in Fig.~\ref{attention_pattern}(c), we counted the attention scores for token pairs on the CodeSearchNet~\cite{DBLP:journals/corr/abs-1909-09436} dataset and found that the attention scores presented a ``long-tail'' phenomenon, i.e. the attention interactions between token pairs are sparse.
By means of top-$k$ sparse attention, calculations of self-attention can be simplified, leading to lower computation overhead.

\begin{figure}[t!]
\includegraphics[width=0.495\textwidth]{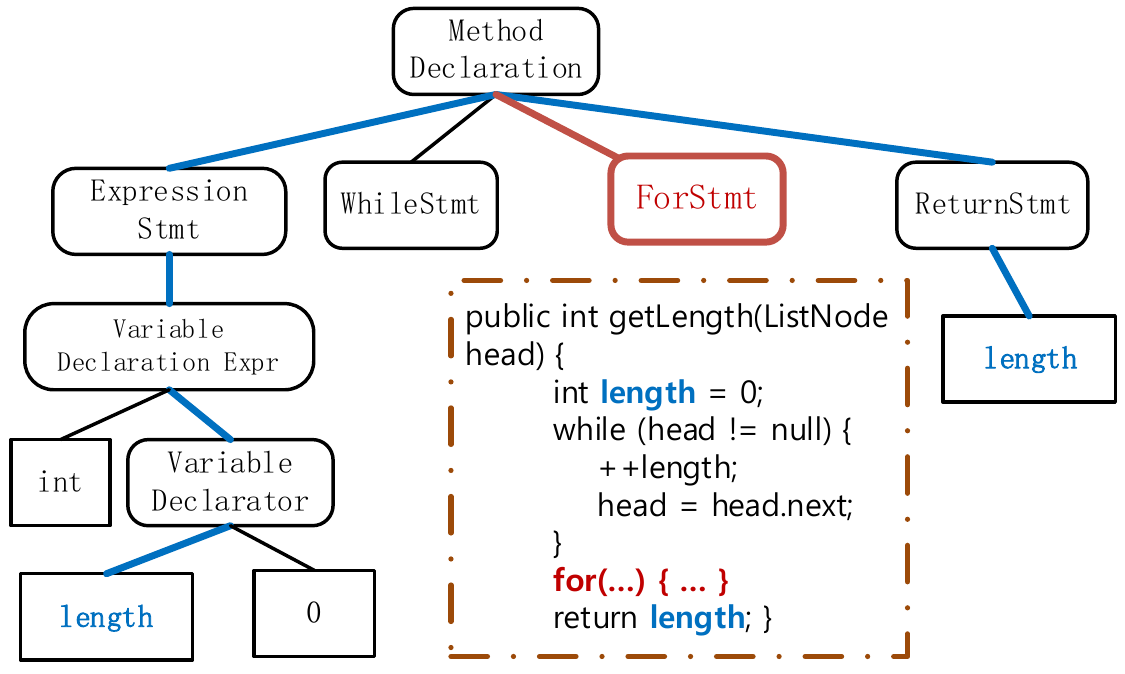}
\caption{An example of the code and its corresponding AST. Some nodes have been omitted for simplicity. The distance between two nodes of ``length'' in AST is not directly related to code length but to the code structure.}
\label{ast_sample}
\end{figure}

Furthermore, the code structure is an important feature in addition to the sequence characteristics of the code. For example,  Abstract Syntax Tree (AST) is a tree that represents the abstract syntactic structure of the source code, leaving out unimportant details such as commas, parentheses. Existing works~\cite{DBLP:conf/icse/ZhangWZ0WL19,DBLP:conf/wcre/WangLM0J20} demonstrate that AST-based models have made significant progress in code-related tasks. Another advantage of AST for long code is that the distance between nodes in AST is not directly related to code length but to the depth of the code structure. 
As shown in Fig.~\ref{ast_sample}, the distance between the two nodes of ``length'' is $6$ (blue path in AST). When the ``For Statement'' is inserted after the ``While Statement'', the distance between these two nodes in the sequence becomes larger, but remains constant in the AST.
Hence, we introduce the AST structures into attention, which explicitly establishes connections to structure-dependent token pairs.

\begin{figure*}[t!]
\includegraphics[width=\textwidth]{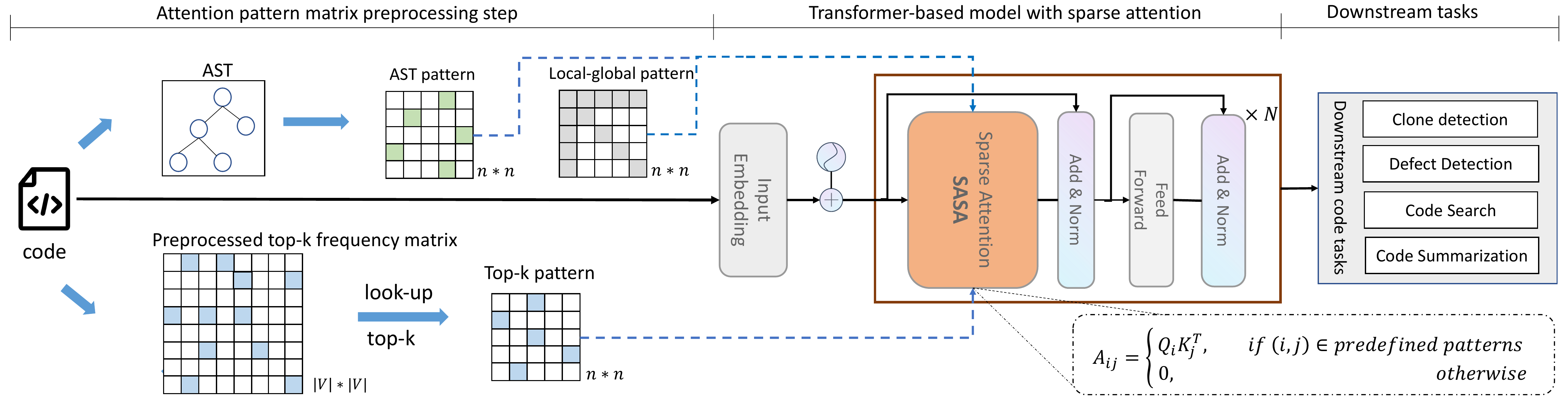}
\caption{Overall architecture based on the Transformer architecture with sparse attention designed to capture context-based and structure-based representations. The matrix of the code's three attention patterns is sparse, with most values being 0. The sparse attention module (SASA) only calculates attention for non-zero positions in the matrix.} 
\label{model_arc}
\end{figure*}

We conducted extensive experiments on four CodeXGLUE~\cite{DBLP:journals/corr/abs-2102-04664} tasks to examine the performance of the proposed SASA model. Results show that SASA outperforms the competing methods in long code tasks. Meanwhile, the SASA-based method also has comparable or better results in full datasets containing mostly short codes. Furthermore, SASA can save $23\%$ to $33\%$ memory with sparse self-attention. 

\section{Related Work}
In this section, we summarize the related works on two aspects: PPLMs and sparse transformers.
\subsection{Program-based Pre-trained Models} Large PPLMs ~\cite{DBLP:journals/corr/abs-2001-00059,DBLP:conf/emnlp/FengGTDFGS0LJZ20,DBLP:journals/corr/abs-2004-13214,DBLP:conf/iclr/GuoRLFT0ZDSFTDC21,DBLP:conf/emnlp/0034WJH21} have significantly improved the performance of code-related downstream tasks. These methods are generally trained on bi-modal data (program languages and natural languages) in a self-supervised manner. SCELMo~\cite{DBLP:journals/corr/abs-2004-13214} trains ELMo~\cite{DBLP:conf/naacl/PetersNIGCLZ18} on corpora of source codes. CuBERT~\cite{DBLP:journals/corr/abs-2001-00059} and CodeBERT~\cite{DBLP:conf/emnlp/FengGTDFGS0LJZ20} follow the architecture of BERT~\cite{DBLP:conf/naacl/DevlinCLT19} and pre-trained by the Masked Language Modeling (MLM) task on CodeSearchNet~\cite{DBLP:journals/corr/abs-1909-09436} corpus, learning the context-aware representations of codes. CodeT5~\cite{DBLP:conf/emnlp/0034WJH21} is a unified pre-trained encoder-decoder model for code understanding and generation tasks. All of these models treat source codes as token sequences and pay little attention to the fact that code structures contain crucial code semantics. Specifically, Abstract Syntax Tree (AST) represents the syntax structure of codes as a tree. Data Flow Graph (DFG) is a graph representation of the transfer between variables. Some task-specific methods~\cite{DBLP:conf/wcre/WangLM0J20,DBLP:conf/acl/WuZZ21,DBLP:conf/icse/ZhangWZ0WL19} use these structures to empower code representations. GraphCodeBERT~\cite{DBLP:conf/iclr/GuoRLFT0ZDSFTDC21} also introduces the edge prediction task to learn representations from the data flow. Most PPLMs are based on the Transformer architecture, and the computational and memory complexity of the self-attention mechanism in the original Transformer is $O(n^2)$ (with $n$ to be the sequence length). Therefore, it is not desirable to directly apply existing PPLMs to long code-related tasks.

\subsection{Sparse Transformer} There have been several studies~\cite{DBLP:journals/corr/abs-1904-10509,DBLP:journals/corr/abs-2004-05150,DBLP:conf/nips/ZaheerGDAAOPRWY20,DBLP:conf/emnlp/QiuMLYW020, DBLP:conf/acl/DaiYYCLS19} that process long sequences in NLP efficiently. BlockBERT~\cite{DBLP:conf/emnlp/QiuMLYW020} divides the attention matrix into $k$ blocks and defines attention on each block, reducing the computational and memory cost to $O(n^2/k)$. Sparse Transformer~\cite{DBLP:journals/corr/abs-1904-10509} and Longformer~\cite{DBLP:journals/corr/abs-2004-05150} employ sliding windows and global tokens to combine local and global information of input sequences. BigBird~\cite{DBLP:conf/nips/ZaheerGDAAOPRWY20} extends random attention on top of Sparse Transformer. These models achieve time and memory savings without significant performance degradation but are not designed for code-related tasks. In this paper, we study processing long codes efficiently and making full use of code structures.

\section{Proposed Framework}

In this section, we elaborate our approach in detail.
Fig.~\ref{model_arc} presents an overview of the proposed framework. We use CodeBERT as the backbone and improve the self-attention to process long and structured code based on three motivations: (1) Due to the high complexity of the self-attention module, CodeBERT cannot handle long codes well with truncated sequences. (2) According to our experimental observations and conclusions of existing works~\cite{DBLP:conf/naacl/BianHCYC21,DBLP:conf/emnlp/KovalevaRRR19}, the calculation of self-attention is redundant. (3) Structure-based representation can complement the sequential semantics of code and improve the performance on downstream tasks~\cite{DBLP:journals/corr/abs-2104-09340,DBLP:conf/acl/WuZZ21}. 
We propose a sparse attention mechanism named SASA, which contains four attention patterns, namely sliding window, global attention, top-$k$ and AST-based attention pattern.

\subsection{Sliding Window and Global Attention}
As shown in Fig.~\ref{attention_pattern}, similar to~\cite{DBLP:conf/emnlp/KovalevaRRR19}, we observe that the attention matrix is very sparse and has some fixed patterns, such as ``vertical'' (for [CLS] or [SEP]) and ``diagonal'' (for local neighbors). Following the BigBird~\cite{DBLP:conf/nips/ZaheerGDAAOPRWY20} architecture, we introduce global tokens and sliding window, corresponding to the vertical and diagonal pattern respectively. 
The computation complexity of the sliding window attention is $O(n \times w)$ with $w<<n$ where $n$ is the sequence length and $w$ is the window size.
Since only a small number of tokens are global tokens, the overall computation complexity of the sliding window and global attention pattern increases linearly with the sequence length.
As modern hardware (such as GPU and TPU) is more suitable for parallel computation, discrete attention computation for sliding window, top-$k$ and AST-aware attention cannot make use of their computational advantages. We follow BigBird~\cite{DBLP:conf/nips/ZaheerGDAAOPRWY20} to divide the attention matrix into blocks and define attention calculations in each block.

Assume that $Q, K \in \mathbb{R}^{n \times d}$ are query and key matrices, where $d$ is the dimension of a token vector. With the block size $b$, we reshape $Q$ and $K$ into $Q'$ and $K'$ with the shape of $\left \lceil n/b \right \rceil \times b \times d$. The attention matrix of sliding window attention pattern is computed as:
\begin{equation}
A^l_{ijst} =
\begin{cases}
\sum_{u}Q'_{isu}K'_{jut}, & \text{if } \left | i-j \right | \leq \left \lfloor \frac{w}{2} \right \rfloor \\
0, & \text{otherwise}
\end{cases}
\label{eq:local}
\end{equation}
where $A^l \in \mathbb{R}^{\left \lceil n/b \right \rceil \times \left \lceil n/b \right \rceil \times b \times b}$, 
$Q'_{isu}$ is the representation of the $i$-th query block and $K'_{jut}$ is the representation of the $j$-th key block. 

For global attention pattern, we specify a set of key blocks as global blocks, denoting as $g$. 
$A^g_{ijst}$ is the attention score of the $i$-th query block and the $j$-th key block. Since every query block attends to global blocks, and global attention is symmetric, the attention calculation is performed when $i$-th query block belongs to $g$ or $j$-th key block belongs to $g$.

\subsection{Top-$k$ Sparse Attention}
As shown in Fig.~\ref{attention_pattern}(a) and Fig.~\ref{attention_pattern}(b), except for the attention pattern of local and global, there are still some scattered parts with high attention scores, which is an effective linguistic feature acquired through pre-training of the model. 
In the trade-off between efficiency and performance, we design a top-$k$ sparse attention pattern that each token attends to the $k$ tokens with the highest attention scores. The computation complexity of this pattern is $O(n \times k)$, which scales linearly with the sequence length. 

To speed up the fine-tuning process, we pre-process the CodeSearchNet corpus and obtain an attention frequency matrix $\widetilde{P}$ of size $\left | V \right | \times \left | V \right |$ where $\left | V \right |$ is the vocabulary size. If the attention score of a token pair is greater than $0.1$ (accounting for more than $10\%$ of the total attention score), the frequency is increased by $1$. A larger value of $\widetilde{P}_{ij}$ means more attention interactions between the $i$-th token and the $j$-th token.
During training, the top-$k$ attention matrix $P \in \mathbb{R}^{n \times n}$ for each input is obtained by looking up the matrix $\widetilde{P}$, and then divided into blocks, denoting as $P' \in \mathbb{R}^{\left \lceil n/b \right \rceil \times \left \lceil n/b \right \rceil \times b \times b}$. The attention frequency of each block is the sum of all the values in the block, that is, add the matrix $P'$ along with the last $2$ dimensions. The attention matrix of top-$k$ attention pattern is computed as:
\begin{equation}
A^p_{ijst} =
\begin{cases}
\sum_{u}Q'_{isu}K'_{jut}, & \text{if } (i,j) \in topk(\sum_{s,t}P'_{ijst}, k) \\
0, & \text{otherwise}
\end{cases}
\label{eq:topk}
\end{equation}
where $A^p \in \mathbb{R}^{\left \lceil n/b \right \rceil \times \left \lceil n/b \right \rceil \times b \times b}$. 
For each query block, we use $topk$ function to select the $k$ key blocks with the highest attention score for calculation.

\begin{figure}
 \renewcommand{\algorithmicrequire}{\textbf{Input:}}
 \renewcommand{\algorithmicensure}{\textbf{Output:}}
 \begin{algorithm}[H]
  \caption{Structure-Aware Sparse Attention (SASA)}
  \label{alg:SASA}
  \begin{algorithmic}[1]
  \REQUIRE $D=\{D_1, \cdots, D_h\}$ where $D_t=\{c_{t,1}, \cdots, c_{t,n}\}$, the train data of each task, $c_t$ is the code tokens of the sample $D_t$; $\widetilde{P} \in \mathbb{R}^{\left | V \right | \times \left | V \right |}, \widetilde{T} \in \mathbb{R}^{\left | V \right | \times \left | V \right |}$, the preprocessed top-$k$ frequency matrix and the AST matrix.
  \ENSURE $A \in \mathbb{R}^{\left \lceil n/b \right \rceil \times \left \lceil n/b \right \rceil \times b \times b}$, the block-based sparse attention matrix.
  \STATE Get the input embedding $X_t$ by the embedding layer of SASA-based model.
  \STATE Reshape the query and key matrix $Q=X_t W_Q, K=X_t W_K \in \mathbb{R}^{n \times d}$ into $Q', K' \in \mathbb{R}^{\left \lceil n/b \right \rceil \times b \times d}$.
  \STATE \textbf{Local pattern} $\leftarrow \{(i, j) | \left | i-j \right | \leq \left \lfloor \frac{w}{2} \right \rfloor$\}.
  \STATE \textbf{Global pattern} $\leftarrow \{(i, j) | i \in g \ or \ j \in g\}$, $g$ is the global blocks.
  
  \STATE \textbf{Top-$k$ pattern}
  \STATE \hspace{1em} Get top-$k$ attention matrix $P \in \mathbb{R}^{n \times n}$ for the input by looking up from the matrix $\widetilde{P}$.
  \STATE \hspace{1em} Divide the matrix $P$ into blocks to get $P' \in \mathbb{R}^{\left \lceil n/b \right \rceil \times \left \lceil n/b \right \rceil}$.
  \STATE \hspace{1em} Top-$k$ pattern $\leftarrow \{(i,j) \in topk(\sum_{s,t}P'_{ijst}, k)$,  $\quad i, j=0, \cdots, \left \lceil n/b \right \rceil-1\}$.
  \STATE \textbf{AST pattern} $\leftarrow \{(i,j) \in topk(\sum_{s,t}T'_{ijst}, k)$.
  \STATE Compute the block-based attention matrix for the input.
  \STATE \hspace{1em} $A_{ijst} \leftarrow \sum_{u}Q'_{isu}K'_{jut}$, if $(i, j)$ belongs to at least one about the above four patterns, otherwise, $A_{ijst} \leftarrow 0$.
    \end{algorithmic}
 \end{algorithm}
\end{figure}

\subsection{AST-aware Structure Attention}
All of the above attention patterns treat codes as sequences.
Besides the sequence characteristics, the structural features of codes are equally important. As shown in Fig.~\ref{ast_sample}, The AST distance represents the structural characteristics between nodes.

Specifically, we parse the code to AST by tree-sitter\footnote{\url{https://tree-sitter.github.io/tree-sitter/.}}. To introduce AST into the attention module, we transform the tree structure into an adjacency matrix. Each non-zero value in the matrix indicates that the corresponding token pair has a structural connection. 
Similar to top-$k$ attention pattern, we divide the adjacency matrix into blocks, resulting in a matrix $T' \in \mathbb{R}^{\left \lceil n/b \right \rceil \times \left \lceil n/b \right \rceil \times b \times b}$. The attention matrix of AST-aware structure attention is computed as:
\begin{equation}
A^t_{ijst} =
\begin{cases}
\sum_{u}Q'_{isu}K'_{jut}, & \text{if } (i,j) \in topk(\sum_{s,t}T'_{ijst}, k) \\
0, & \text{otherwise}
\end{cases}
\label{eq:ast}
\end{equation}
where $A^t \in \mathbb{R}^{\left \lceil n/b \right \rceil \times \left \lceil n/b \right \rceil \times b \times b}$. 

\subsection{Model Summary}
Based on the CodeBERT pre-trained model, we replace self-attention with SASA and fine-tune it on downstream tasks. We have four patterns in SASA, and each pattern has an attention matrix. 
As shown in Algorithm \ref{alg:SASA}, each query block in SASA attends to $w$ sliding window blocks, $g$ global blocks, $k$ top-$k$ blocks and AST blocks with the total cost of $O(n(w+g+k)b)$. 

\begin{table*}
\centering
\caption{Main results for long codes. The best results are in bold font, and the second best are underlined.}
\begin{tabular}{l | c | c | c | c | c | c}
 \toprule 
\multirow{2}{*}{\bf Method} & \multicolumn{3}{c|}{\bf BigCloneBench} & \bf Defect Detection & \bf Code Search & \bf Code Summarization \\
\cline{2-7}
~ & \bf Precision & \bf Recall & \bf F1 Score & \bf Accuracy (\%) & \bf MRR & \bf BLEU-4 \\
 \hline
Roberta-base & 0.613 & 0.831 & 0.706 & 52.59 & 25.60 & 9.17 \\
CodeBERT & 0.718 & 0.874 & 0.789 & 55.71 & 38.37 & 12.26 \\
GraphCodeBERT & 0.852 & 0.891 & 0.871 & 55.39 & 43.13 & \textbf{12.93} \\
Longformer & 0.752 & \underline{0.901} & 0.820 & \underline{56.90} & \underline{44.09} & 12.20 \\
BigBird & \underline{0.892} & 0.898 & \underline{0.895} & 56.79 & 41.89 & 12.42 \\
\hline
\bf SASA & \textbf{0.906} & \textbf{0.917} & \textbf{0.911} & \textbf{57.44} & \textbf{46.36} & \underline{12.88} \\
\bottomrule
\end{tabular}
\label{tab:main_exp}
\end{table*}

\section{Experiments}

In this section, we conduct extensive experiments to evaluate the proposed SASA framework.

\subsection{Experimental Settings}
\subsubsection{Tasks} 
To evaluate the effectiveness of our proposed SASA model, we provide results and ablations on four code understanding tasks as follows: (1) Clone detection is a binary classification task for code pairs, where $1$ stands for semantic equivalence and $0$ for others. (2) Defect detection (Defect for short) task aims to identify whether a program may attack a software system. (3) Code search aims to search source code that matches the input natural language. (4) Code summarization (Summ for short) task is to generate natural language comments for a given code. We use the datasets provided by CodeXGLUE~\cite{DBLP:journals/corr/abs-2102-04664} and select the code with the length greater than 1024 as the long datasets.

\subsubsection{Baselines} 

We compare SASA with several pre-trained models. Roberta~\cite{DBLP:journals/corr/abs-1907-11692} is a multi-layer
bidirectional transformer encoder.
CodeBERT~\cite{DBLP:conf/emnlp/FengGTDFGS0LJZ20} uses Roberta~\cite{DBLP:journals/corr/abs-1907-11692} as the backbone and is continuously pre-trained on codes from CodeSearchNet~\cite{DBLP:journals/corr/abs-1909-09436}.
GraphCodeBERT~\cite{DBLP:conf/iclr/GuoRLFT0ZDSFTDC21} follows BERT~\cite{DBLP:conf/naacl/DevlinCLT19} architecture and is pre-trained on CodeSearchNet corpus with an edge prediction task. Longformer~\cite{DBLP:journals/corr/abs-2004-05150} and Bigbird~\cite{DBLP:conf/nips/ZaheerGDAAOPRWY20} is a sparse attention model with local and global attention patterns that works well with long sequences in NLP.

\subsubsection{Model Configuration} 

We follow CodeBERT~\cite{DBLP:conf/emnlp/FengGTDFGS0LJZ20} as the model backbone, the difference being that we set the sequence length to 1024 instead of 512. In addition, we set $w=3$, $\left | g \right |=2$, $k=3$, $b=32$ for all experiments. The other hyperparameters follow the same setting provided by CodeXGLUE~\cite{DBLP:journals/corr/abs-2102-04664}. 


\subsection{Overall Performance}
Table ~\ref{tab:main_exp} shows the overall performance on four long code datasets. We can see that: (1) The performance of models that truncate long sequences, such as Roberta-base, CodeBERT, GraphCodeBERT, is greatly degraded. (2) GraphCodeBERT works better than CodeBERT, suggesting that the structural properties of code are important in understanding long codes. (3) Sparse attention models that are effective for long sequences in NLP cannot be directly used for long code tasks, but Longformer and BigBird outperform CodeBERT by loading the pre-training weights of CodeBERT. (4) The SASA model exploits both the advantages of sparse attention mechanism and code structures, and has competitive performance compared with state-of-the-art models.

\subsection{Ablation Study}
We conduct an ablation study on two tasks to examine the effect of top-$k$ and AST attentions, as shown in Table~\ref{tab:analysis}. Without top-$k$ attention pattern, the F1 score of BigCloneBench (BCB for short) decreases by $0.083$. The accuracy of Defect Detection decreases by $0.6\%$. This indicates that for each query block, $k$ key blocks with the highest attention scores can already cover most of the interactive information. Meanwhile, compared to Longformer and BigBird, it is useful to explicitly introduce the structure connection of codes into attention.

\subsection{Memory Analysis}

\begin{figure}
\centering
\includegraphics[width=0.425\textwidth]{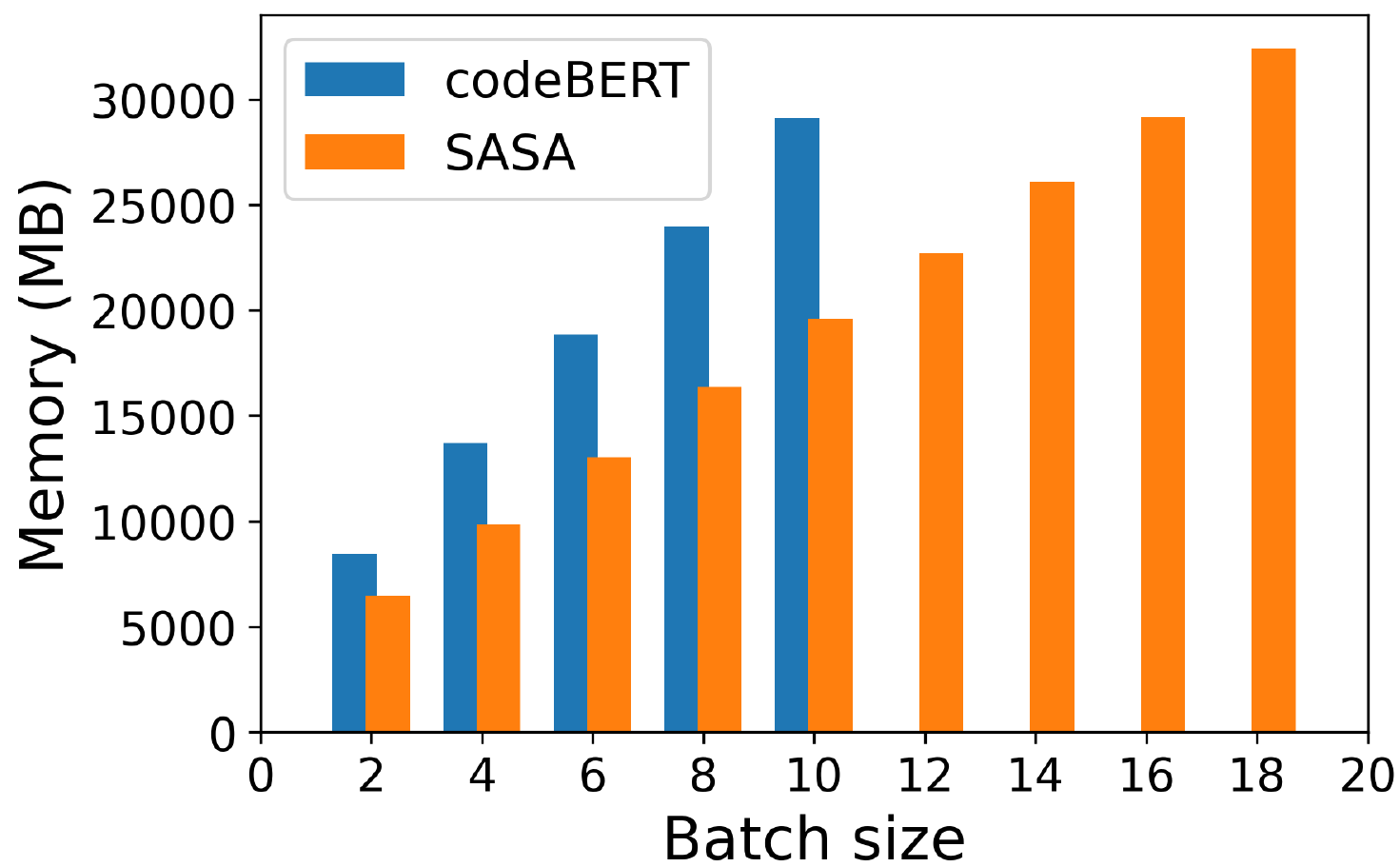}
\caption{Memory used by the model. The memory consumption of larger batch sizes for codeBERT is not shown due to out-of-memory (OOM) error.}
\label{fig:memory}
\end{figure}

The sparse attention mechanism reduces computation and memory complexity compared to full self-attention. Fig.~\ref{fig:memory} shows the memory usage  of CodeBERT and SASA. When batch size is 8, SASA saves more than 7GB of memory compared to CodeBERT. When batch size is greater than $10$, CodeBERT has an out-of-memory problem on a 32GB V100 machine.

\begin{table}
\centering
\caption{Ablation study of SASA.}
\begin{tabular}{@{}l|l|l|l|c}
 \toprule 
\multirow{2}{*}{\bf Method} & \multicolumn{3}{c|}{\bf BigCloneBench} & \bf Defect Detection \\
 \cline{2-5}
~ & \bf Pre. & \bf Rec. & \bf F1 & \bf Accuracy (\%) \\
\hline
\bf SASA & \bf 0.906 & 0.917 & \bf 0.911 & \bf 57.44 \\
 w/o. top-$k$ & 0.753 & 0.919 & 0.828 & 56.84 \\
 w/o. AST & 0.830 & \bf 0.922 & 0.874 & 57.22 \\
\bottomrule
\end{tabular}
\label{tab:analysis}
\end{table}

\section{Conclusion}
In this paper, we present SASA, a Structure-Aware Sparse Attention mechanism, which reduces the complexity and improves performance for long code understanding tasks. SASA consists of two novel components, which are top-$k$ sparse attention and Abstract Syntax Tree (AST)-based structure-aware attention. The former simplifies the complexity of self-attention while capturing the rich sequential context. As a complement, AST attention explicitly builds the structural link between tokens. SASA achieves performance improvements and resource-saving in multiple tasks.
In industrial scenarios, SASA can be used to generate code comments and retrieve code based on natural language queries.


\section*{Acknowledgments}
\label{sec:acknowledge}
This work has been supported by the National Natural Science Foundation of China under Grant Nos. U1911203 and 61877018, and Alibaba Group through the Alibaba Innovation Research Program.


\balance

\end{document}